\documentclass{article}
\usepackage{spconf,amsmath,graphicx}
\usepackage{xcolor, soul}
\sethlcolor{pink}
\usepackage{enumitem}
\usepackage[normalem]{ulem}
\usepackage{booktabs}
\usepackage{url}

\title{SPEAKER TAGGING CORRECTION WITH NON-AUTOREGRESSIVE LANGUAGE MODELS}
\name{Grigor Kirakosyan$^{1,2}$, Davit Karamyan$^{2,3}$}
\address{
  $^1$Institute of Mathematics of NAS RA, Yerevan\\
  $^2$Krisp.ai, Yerevan \\
  $^3$ML Lab at YSU, Yerevan
}
\begin{document}
%
\maketitle
\begin{abstract}
Speech applications dealing with conversations require not only recognizing the spoken words but also determining who spoke when. The task of assigning words to speakers is typically addressed by merging the outputs of two separate systems, namely, an automatic speech recognition (ASR) system and a speaker diarization (SD) system. In practical settings, speaker diarization systems can experience significant degradation in performance due to a variety of factors, including uniform segmentation with a high temporal resolution, inaccurate word timestamps, incorrect clustering and estimation of speaker numbers, as well as background noise. 

Therefore, it is important to automatically detect errors and make corrections if possible. We used a second-pass speaker tagging correction system based on a non-autoregressive language model to correct mistakes in words placed at the borders of sentences spoken by different speakers. We first show that the employed error correction approach leads to reductions in word diarization error rate (WDER) on two datasets: TAL and test set of Fisher. Additionally, we evaluated our system in the Post-ASR Speaker Tagging Correction challenge and observed significant improvements in cpWER compared to baseline methods.

\end{abstract}

\section{Introduction}
\label{sec:intro}
Speech recognition systems have advanced significantly in the past decade. Still, even with these remarkable advances, machines have difficulties understanding natural conversations with multiple speakers, such as in broadcast interviews, meetings, telephone calls, videos or medical recordings. One of the first steps in understanding natural conversations is to recognize the words spoken and their corresponding speakers. SD determines "who spoke when" in multi-speaker audio and is a crucial part of the speech translation system. SD is used in conjunction with ASR to assign a speaker label to each transcribed word and has widespread applications in generating meeting/interview transcripts, medical notes, automated subtitling and dubbing, downstream speaker analytics, among others. Usually, this is done in multiple steps that include (1) transcribing the words using an ASR system, (2) predicting "who spoke when" using a speaker diarization system, and, finally, (3) reconciling the output of those two systems \cite{paturi2023lexical}. A typical reconciliation algorithm works as follows: (1) If the word segment overlaps with at least one speaker segment, then this word is associated with the speaker that has the biggest temporal overlap with this word; (2) otherwise, if this word segment does not overlap with any speaker segment, then it is associated with the speaker that has the smallest temporal distance to this word based on the segment boundaries \cite{wang2024diarizationlm}.

Speaker diarization systems often face numerous challenges that can lead to subpar performance, negatively impacting the user's perception of transcript quality. However, some of these errors can be mitigated through post-correction techniques. In this work, we first analyze the mistakes made during reconciliation and categorize them. We then implement a speaker error correction module to rectify inaccuracies, particularly for boundary words between sentences spoken by different speakers.

\begin{table*}[ht]
\centering
\caption{Examples of different diarization errors (errors are underlined and marked in pink color).}
\label{t:errors}
\begin{tabular}{c|l}
\multicolumn{1}{l|}{Error Type} & Example                                                                                                              \\ \hline
Type a                          & \begin{tabular}[c]{@{}l@{}}\textbf{Speaker A:}\\ right that's going exactly going back to facebook's optimizer algorithm that's not optimizing for truth right \\ it's optimizing for profit and they they claim to be neutral but of course nothing's neutral \underline{\hl{right}} and we have \\ seen the results we've seen what it's actually optimized for and it's not pretty\end{tabular}                                                                                             \\ \hline
Type b                          & \begin{tabular}[c]{@{}l@{}}\textbf{Speaker A:} \\ and presumably you could take all that biased input data and say this high chance recidivism means that we \\ should rehabilitate more i mean like you could take all that same stuff and choose to do a completely different \\ thing with the result of\\ \textbf{Speaker B:} \\ \underline{\hl{the algorithm}} total that's exactly my point exactly my point you know we could say oh i wonder why people \\ who have this characteristic have so much worse recidivism well let's try to help them find a job maybe that'll \\ help we could use those algorithms those risk scores to try to account for our society\end{tabular} \\ \hline
Type c                          & \begin{tabular}[c]{@{}l@{}}\textbf{Speaker A:} \\ \underline{\hl{is this a good or bad thing that social media has been able to infiltrate politics}}\end{tabular}                                                                                                 
\end{tabular}
\end{table*}

\section{RELATED WORK}
\label{sec:relwork}

We are inspired by the work in \cite{paturi2023lexical}, where the authors introduced the speaker error corrector (SEC). SEC corrects speaker errors at the word level without modifying the underlying ASR or acoustic SD systems. In \cite{paturi2023lexical}, word embeddings from the ASR transcript are extracted using a pre-trained RoBERTa-base language model (LM) \cite{liu2019roberta}. These embeddings, along with the hypothesized speaker labels, are fed into a separately trained transformer encoder, which produces the corrected speaker labels. The transformer encoder is trained on both simulated diarization errors and real data.

In \cite{wang2024diarizationlm}, the authors proposed DiarizationLM, a framework to leverage large language models (LLM) to post-process the outputs from a speaker diarization system. In this framework, the outputs of the ASR and SD systems are represented in a compact textual format and included in the prompt to an optionally finetuned LLM. The outputs of the LLM can be used as the refined diarization results with the desired enhancement. As a post-processing step, this framework can be easily applied to any off-the-shelf ASR and speaker diarization systems without retraining existing components.

More recently, in \cite{park2024enhancing}, the authors suggested using LLM to predict the speaker probability for the next word and incorporating this probability into the beam search decoding of speaker diarization. In this approach, prompting is implemented word-by-word, unlike in \cite{wang2024diarizationlm}, where a single prompt is used to post-process the entire speaker diarization results. One drawback of this proposed approach is that it requires word-level speaker probabilities for beam search decoding, which may be absent in some SD systems.

\section{CLASSIFYING DIARIZATION ERRORS}
\label{sec:errors}

A typical method for assessing traditional speaker diarization systems is the diarization error rate (DER). This is calculated by adding together three types of errors: false alarms, missed detections, and speaker confusion errors. Essentially, DER compares the reference speaker labels with the predicted speaker segments in the time domain. On the other hand, the use of a joint ASR and SD system directly assign speakers to recognized words, eliminating the need to rely on time boundaries. In \cite{Shafey2019JointSR}, the authors proposed a new metric, word diarization error rate (WDER), to evaluate such joint ASR and SD systems, by measuring the percentage of words in the transcript that are tagged with the wrong speaker:
$$
WDER = \frac{S_{IS}+C_{IS}}{S+C}
$$
where $S_{IS}$ represents the number of ASR substitutions with incorrect speaker tags, $C_{IS}$ represents the number of correctly recognized ASR words with incorrect speaker tags, S is the total number of ASR substitutions and C is the total number of correctly recognized ASR words. WDER doesn't take into account deletion and insertion errors as the speaker tags associated with them cannot be mapped to reference without ambiguity.

One benefit of WDER is that it can be used to automatically identify and visualize diarization errors at the word level. By examining errors at the word level, it is possible to categorize them into three categories:

\begin{enumerate}[label=(\alph*)]
    \item Incorrect speaker tags within a paragraph
    \item The first and last words of a paragraph having incorrect speaker tags
    \item A complete paragraph being assigned to the wrong speaker
\end{enumerate}

The main cause of errors of type (a) and (b) is the use of uniform audio segmentation with a high temporal resolution. Inaccurate  word timestamps can also lead to type (b) errors. Type (c) errors typically occur due to inaccurate estimation of the number of speakers and incorrect clustering. Background noise, music and reverberation also contribute to all types of errors. Examples of each type of error are illustrated in Table \ref{t:errors}.

\section{SPEAKER ERROR CORRECTOR}
\subsection{System overview}
\label{ssec:sec}

We use the lexical speaker error corrector introduced in \cite{paturi2023lexical}, which aims to improve diarization accuracy by leveraging lexical information. In this approach, word embeddings are extracted using a pre-trained RoBERTa-base LM. These embeddings, along with the hypothesized speaker labels, are fed into a transformer encoder, which produces the corrected speaker labels.

In contrast to the original implementation, we use an ALBERT-base LM \cite{lan2019albert} due to its memory efficiency. Additionally, our error simulation procedure differs from the original work, where the target words are substituted with random words. We have also replaced the standard cross-entropy loss with a permutation invariant loss. The next section will cover all the training details.

\subsection{Training details}
\label{ssec:training}
We train the SEC on two-speaker scenarios, generating synthetic errors for both words and speaker tags. For word errors, we employ an alternative spelling prediction (ASP) model \cite{fox22_interspeech}. It aims to predict how the ASR system might inaccurately recognize a given word without executing the ASR model itself. For speaker tag errors, we simulate errors at speaker change points if the input involves two speakers, as shown in Example 1. If the input contains only one speaker, we simulate errors only at the beginning or at the end of the input, as illustrated in Example 2.

\begin{itemize}[leftmargin=*]
    \item \textbf{Example 1:}
    
    \textbf{Reference}: $<$spk1$>$ can you study with the radio on $<$spk2$>$ no i listen to background music \\
    \textbf{Simulated}: $<$spk1$>$ can you study with the radio on no i $<$spk2$>$ listen to background music
    
    \item \textbf{Example 2:}
    \item[] \textbf{Reference}: $<$spk1$>$ uh huh it but it almost makes me feel like \\
    \textbf{Simulated}: $<$spk1$>$ uh huh it but it almost makes me feel $<$spk2$>$ like
    
\end{itemize}

Our goal is to accurately predict speaker segmentation, even though the concept of speaker ID can sometimes be ambiguous. Consider the motivating example illustrated in Table \ref{table: ex}. The model can either correct the first two tags or the last five tags. To handle such cases, we use permutation invariant cross-entropy loss for speaker tag classification, which selects a permutation of speakers that results in the minimum loss.

\begin{table}[]
\centering

\caption{Example of ambiguous sample.}
\begin{tabular}{@{}l|l@{}}
\toprule
Reference Tags       & spk1 spk1 spk1 spk1 spk1 spk1 spk1 \\ \midrule
Simulated Tags & spk1 spk1 spk2 spk2 spk2 spk2 spk2 \\ \bottomrule
\end{tabular}
\label{table: ex}
\end{table}

\subsection{Inference setup}
\label{ssec:infer}
During inference, we perform error correction only at speaker change points. We define a context window around these change points and feed the window, along with the hypothesized speaker tags, into a SEC model. The window consists of up to 18 words from the left context and 18 words from the right context, up to the nearest change points.

\begin{table}[t]

\centering
\resizebox{0.45\textwidth}{!}
{
\begin{tabular}{@{}rccc@{}}
\toprule
\multicolumn{1}{c}{\textbf{Word}} & \textbf{Top1} & \textbf{Top2} & \textbf{Top3} \\ \midrule
\multicolumn{4}{c}{\textbf{Words seen during training}}                          \\ \midrule
\multicolumn{1}{r|}{hashimoto}       & hashamoto     & hashimoto     & hashamato     \\
\multicolumn{1}{r|}{jupyter}         & jupiter       & jupitor       & jupitter      \\
\multicolumn{1}{r|}{kotlin}          & cotlin        & cotlan        & codlin        \\
\multicolumn{1}{r|}{pulumi}          & pulumi        & polumi        & poulumi       \\ \midrule
\multicolumn{4}{c}{\textbf{Words unseen during training}}                        \\ \midrule
\multicolumn{1}{r|}{farnoosh}        & farnosh       & farnush       & farnash       \\
\multicolumn{1}{r|}{doernenburg}     & dornenburg    & doernenberg   & doernenburg   \\
\multicolumn{1}{r|}{odersky}         & oderski       & odersky       & odderski      \\ \bottomrule
\end{tabular}

}
\caption{Top three alternates generated by the ASP model.}
\label{table: asp_examples}
\end{table}

\section{Results}

\begin{table}
\centering

\begin{tabular}{l c} 
\textbf{Model} & \textbf{BLEU} \\ [0.2ex] 
\hline 
{Identity}  & {0.48}\\ [0.2ex]
{ASP with greedy decoding}  & {0.6}\\ [0.2ex]
{ASP with beam search}  & {0.605} \\
\hline 
\end{tabular}
\caption{Performance of the ASP model with and without beam search in comparison to the identity baseline.}
\label{table:bleu} 
\end{table}

\begin{table*}[ht]
\centering
\caption{Example case from the TAL testing set (errors are underlined and marked in pink color).}
\label{table:sec-example}
\begin{tabular}{@{}ll@{}}
\toprule
\multicolumn{1}{c}{Before Correction}                                                                                                                                                                      & \multicolumn{1}{c}{After Correction}                                                                                                                                                                      \\ \midrule
\begin{tabular}[c]{@{}l@{}}{[}spk2{]}: three percent to five percent  you mean of all \\ healthcare\\ \\ {[}spk5{]}: \underline{\hl{professionals}} all across the profession \underline{\hl{wow}}\\ \\ {[}spk2{]}: which drugs\end{tabular} & \begin{tabular}[c]{@{}l@{}}{[}spk2{]}: three percent to five percent you mean of all \\ healthcare professionals\\ \\ {[}spk5{]}: all across the profession\\ \\ {[}spk2{]}: wow which drugs\\
\end{tabular} \\ \bottomrule
\end{tabular}
\end{table*}

\subsection{Evaluation setup}
\label{ssec:eval}
In this work, we use the full Fisher \cite{cieri2004fisher, cieri2005fisher}, DailyDialog \cite{li2017dailydialog}, and SLT GenSEC Challenge Track-2 training datasets to train the SEC model. We split the Fisher data into training, validation and test sets as defined in \cite{wang2022highly}. For evaluation, in addition to the Fisher test split, we use the standard test split of the TAL \cite{Mao2020SpeechRA} dataset. For internal evaluations and model selection, we report performance using the WDER, as we believe it provides a more accurate representation of a speaker diarization system's performance at the word level compared to the cpWER metric \cite{Povey2011TheKS, MeetEval23}. Our final evaluation for the Post-ASR Speaker Tagging Correction challenge is conducted using the cpWER metric.

\par We use the pre-trained FastConformer-Large model\footnote{\url{https://catalog.ngc.nvidia.com/orgs/nvidia/teams/nemo/models/stt_en_fastconformer_ctc_large}} \cite{rekesh2023fast} to transcribe test datasets and then diarize them using the Titanet-Small\footnote{\url{https://catalog.ngc.nvidia.com/orgs/nvidia/teams/nemo/models/titanet_small}} \cite{koluguri2022titanet} embedding extractor along with the standard spectral clustering \cite{von2007tutorial}. After generating the transcript, we apply our SEC model to it and compare the corrected speaker tags with the ground truth tags using either WDER or cpWER.

\subsection{Alternate spelling prediction model}
\label{ssec:asp}

To train the alternate spelling prediction model, we use roughly 1.15 million word pairs that were mistakenly recognized by a Conformer-medium model \cite{gulati20_interspeech}. We use a medium-size pre-trained Conformer checkpoint\footnote{\url{https://catalog.ngc.nvidia.com/orgs/nvidia/teams/nemo/models/stt_en_conformer_ctc_medium}} that was made available by Nvidia. Furthermore, we removed error pairs in which the phonetic forms of the reference and predicted words had an edit distance greater than 50\%. We used the grapheme-to-phoneme (G2P)  library\footnote{\url{https://github.com/Kyubyong/g2p}} to convert words to the corresponding phoneme sequence.

Similar to \cite{fox22_interspeech}, our ASP model is also based on a transformer encoder-decoder framework. It has two layers in both the encoder and decoder with two attention heads per layer and 400 units per layer resulting in a total of 6.5M parameters. However, unlike the original paper, the input and the output subword tokenization is the same as the tokenization used for the ASR model. 

At inference time, we use beam search to produce a 3-best list of alternate spellings for each word. During the training of the SEC model, we generate ASR errors by replacing the target word with a randomly picked alternate. 

To test the accuracy of the ASP model, we measure the BLEU score \cite{papineni2002bleu} between the word pieces of the reference and predicted alternates. Table \ref{table:bleu} shows the results of the ASP model on the test set. For comparison, we present the baseline score for an identity system that keeps the input word unchanged. In addition, we report the score obtained by a refined ASP model using beam search. Table \ref{table: asp_examples} illustrates examples of alternates that the ASP model produces.

\subsection{SEC system}
\label{ssec:sec-sys}
We use a pre-trained ALBERT-base model as the backbone LM and a transformer encoder with 128 hidden states. For word error simulation, we either leave the word unchanged or substitute it with a corresponding alternate generated by the ASP model with a probability of 0.1. For speaker errors, we introduce a maximum of two errors: in 40\% of inputs, no errors are simulated; in 48\% of inputs, a single speaker tag error is generated; and in 12\% of inputs, two speaker tag errors are simulated. The model is trained with an average sequence length of 30 words per batch, which was found to be optimal through hyperparameter search in \cite{paturi2023lexical}. Initially, we trained only the transformer encoder part of the SEC model. Subsequently, we unfreeze the ALBERT part and train the entire SEC model.

\subsection{Results}
\label{ssec:results}

\begin{table}[t]
\caption{The performance of the SEC model on the Fisher and TAL datasets. The results are reported in WDER. The x/y notation signifies the number of incorrectly assigned words (x) out of the total words analyzed (y).}
\begin{tabular}{@{}lcc@{}}
\toprule
Model Type    & Fisher               & TAL                   \\ \midrule
No Correction & 2.8 (7673/274398)  & 4.25 (14487/340991) \\
SEC           & 2.42 (6653/274398) & 4.11 (14012/340991) \\ \bottomrule
\end{tabular}
\label{table WDER}
\end{table}

\begin{table}[t]

\caption{The performance of the SEC model on the SLT GenSEC Challenge Track-2 dev and eval datasets. The results are reported in cpWER.}
\begin{tabular}{@{}lcc@{}}
\toprule
Model Type          & dev                                        & eval                                       \\ \midrule
No Correction       & 24.64 (5998/24335)                         & 28.45 (5563/19552)                         \\
Baseline & 24.53 (5971/24335) & 28.36 (5546/19552) \\
SEC                 & 23.97 (5834/24335)                         & 27.76 (5429/19552)                         \\ \bottomrule
\end{tabular}
\label{table cpWER}
\end{table}

From Table \ref{table WDER}  and Table \ref{table cpWER} we can see that the SEC model consistently outperforms the "No Correction" baseline across different datasets. For instance, on the Fisher dataset, the SEC model reduces the WDER from 2.8\% to 2.42\%. Similarly, on the TAL dataset, the WDER decreases from 4.25\% to 4.11\%. This improvement is also evident in the cpWER metric for the SLT GenSEC Challenge Track-2 datasets, where the SEC model achieves lower error rates on both the dev and eval sets compared to the baseline \cite{park2024enhancing}. Table \ref{table:sec-example} presents an example case from the TAL testing set, where we see improvements after applying the speaker error correction model.

One drawback of our method is that it is only applied to speaker change points. When we apply the model more frequently, we observe performance degradation. 

\section{CONCLUSIONS}
\label{sec:concl}

In this work, we implemented a speaker error correction model to correct word-level speaker errors for boundary words between sentences spoken by different speakers. We achieve this using a language model over the ASR transcriptions to correct the speaker labels. We train the SEC model using only text data by simulating both word errors and speaker errors without the need for any paired audio-text data. For simulating word errors, we train an alternate spelling prediction model that can predict how the ASR will recognize a given word. We achieved an absolute reduction in WDER of over 0.38\% and 0.14\% across the Fisher test and TAL datasets, respectively.  Additionally, we evaluated our system in the Post-ASR Speaker Tagging Correction challenge and observed  improvements in cpWER compared to baseline methods. The proposed SEC framework is also lightweight and can be integrated as a post-processing module over existing on-device ASR-SD systems.

\end{document}